\documentclass[sigconf]{acmart}

\usepackage{graphicx}
\usepackage{balance}  
\usepackage{amsfonts}
\usepackage{amssymb}
\usepackage{color,soul}
\usepackage{hyperref}
\usepackage{algorithm}
\usepackage{algorithmic}
\usepackage{subcaption}
\usepackage{changes}
\usepackage{comment}
\usepackage{xspace}
\usepackage{boxedminipage}
\usepackage{diagbox}
\usepackage{multirow}
\usepackage{paralist}
\usepackage{amsmath}
\usepackage{enumitem}
\usepackage{array}
\usepackage{mdframed}
\usepackage[normalem]{ulem}
\usepackage{tabularx}
\usepackage{pgf,tikz}
\usepackage{pgfplots}
\usepackage{mathrsfs}
\usepackage{diagbox}

\newcolumntype{L}[1]{>{\raggedright\let\newline\\\arraybackslash\hspace{0pt}}m{#1}}
\newcolumntype{X}[1]{>{\centering\let\newline\\\arraybackslash\hspace{0pt}}p{#1}}
\newcommand{\algname}{\emph{MLAT}}

\AtBeginDocument{%
  \providecommand\BibTeX{{%
    \normalfont B\kern-0.5em{\scshape i\kern-0.25em b}\kern-0.8em\TeX}}}

\copyrightyear{2019}
\acmYear{2018}
\setcopyright{acmlicensed}
\acmConference[MileTS '19]{MileTS '19: 5th KDD Workshop on Mining and Learning from Time Series}{August 5th, 2019}{Anchorage, Alaska, USA}
\acmBooktitle{MileTS '19: 5th KDD Workshop on Mining and Learning from Time Series, August 5th, 2019, Anchorage, Alaska, USA}
\acmPrice{15.00}
\acmDOI{10.1145/1122445.1122456}
\acmISBN{978-1-4503-9999-9/18/06}

\begin{document}

%
\title{MLAT: Metric Learning for kNN in Streaming Time Series}

%
\author{Dongmin Park, Susik Yoon, Hwanjun Song, Jae-Gil Lee}
\authornote{Jae-Gil Lee is the corresponding author.}
\affiliation{%
  \institution{Graduate School of Knowledge Service Engineering, KAIST}
}
\email{{dongminpark, susikyoon, songhwanjun, jaegil}@kaist.ac.kr}

%


%
\begin{abstract}
Learning a good distance measure for distance-based classification in time series leads to significant performance improvement in many tasks. Specifically, it is critical to effectively deal with \emph{variations} and \emph{temporal dependencies} in time series. However, existing metric learning approaches focus on tackling variations mainly using a strict alignment of two sequences, thereby being not able to capture temporal dependencies. To overcome this limitation, we propose \emph{\textbf{\algname{}}}, which covers both alignment and temporal dependencies at the same time. \algname{} achieves the alignment effect as well as preserves temporal dependencies by augmenting a given time series using a sliding window. Furthermore, \algname{} employs time-invariant metric learning to derive the most appropriate distance measure from the augmented samples which can also capture the temporal dependencies among them well. We show that \algname{} outperforms other existing algorithms in the extensive experiments on various real-world data sets.
\end{abstract}

\ccsdesc[500]{Mathematics of computing~Time series analysis}
\ccsdesc[300]{Theory of computation~Machine learning theory}
\ccsdesc[100]{Computing methodologies~Supervised learning by classification}
%
\keywords{Streaming Time Series, Metric Learning, Optimization}

%

\maketitle

\section{Introduction}
\label{sec:introduction}

Streaming time series classifications play an increasingly important role in activity recognition\,\cite{Mueen16} and fraud detection\,\cite{Fortuny14}. 
However, since the number of labels in streaming time series data is often insufficient to build a high-quality classifier\,\cite{Wei06}, $k$-nearest neighbor (kNN), a non-parametric method, is widely used and it empirically results in high accuracy in several applications\,\cite{Ding08}.

\begin{figure}[t!]
\centering
\begin{subfigure}[t]{0.48\textwidth}
\centering
\hspace*{-0.1cm}
\includegraphics[width=0.85\textwidth]{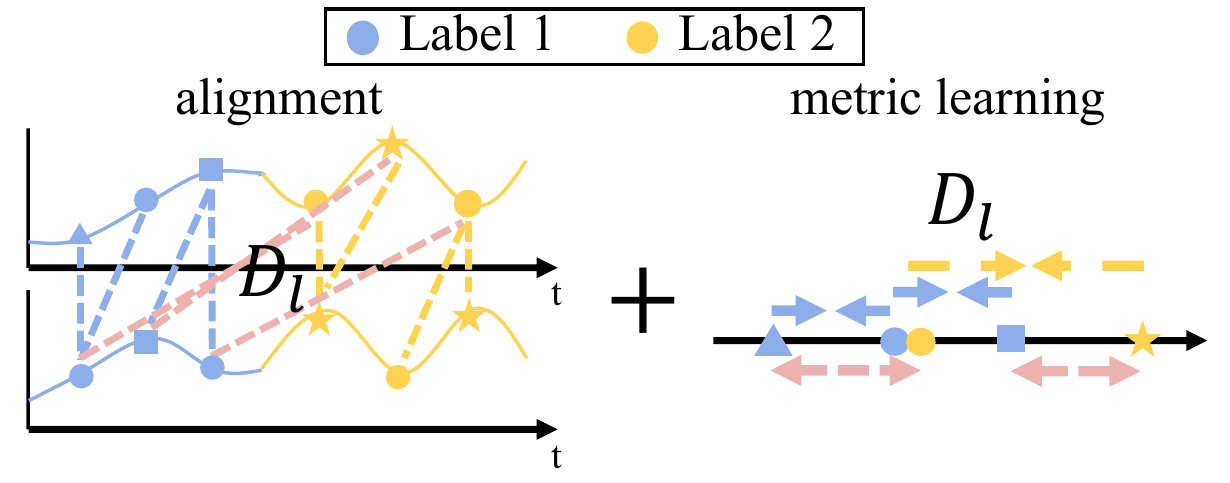} 
\vspace*{-0.3cm}
\caption{Metric learning on local distances.}
\vspace*{-0.05cm}
\label{fig:A}
\end{subfigure}
\begin{subfigure}[b]{0.48\textwidth}
\centering
\includegraphics[width=0.85\textwidth]{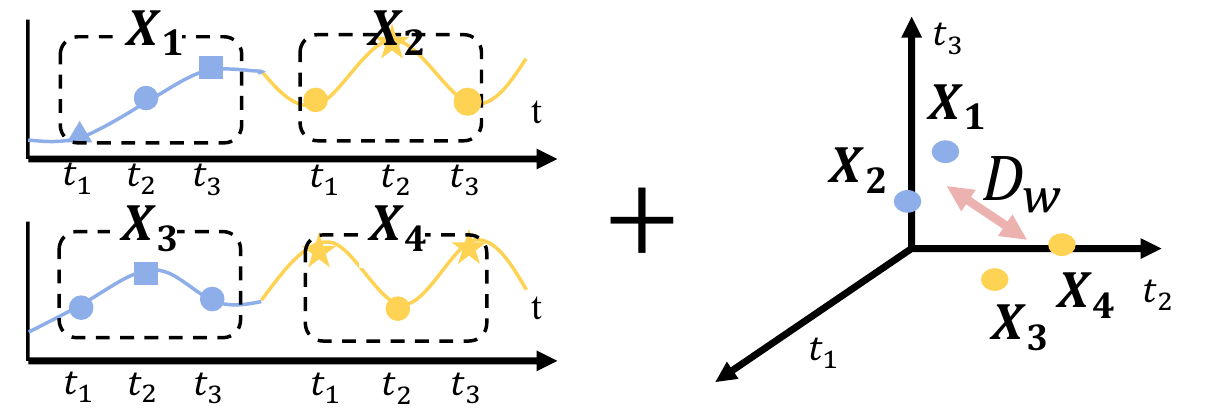} 
\vspace*{-0.2cm}
\caption{Metric learning on window distances.}
\vspace*{-0.4cm}
\label{fig:B}
\end{subfigure}    
\caption{The effect of alignment to temporal dependency.}
\vspace*{-0.7cm}
\label{fig:motivation}
\end{figure}

In accordance with this trend, 
many studies have been focused on improving the performance of kNN by designing more appropriate distance measures for a given data set.
Recently, various time series metric learning approaches\,\cite{Mei15, Shen17, Che17} have been developed to achieve this goal. 
Because variations in time series data, such as sequence shifting and scaling, is one of the main challenges, the existing algorithms consist of the two phases: \emph{(i) alignment} and \emph{(ii) metric learning}. As shown in Figure \ref{fig:motivation}(a), the subsequences of the time series data are aligned in pairs to match the optimal time steps\,(dotted line), rendering them robust to variation.
Metric learning is subsequently conducted on the \emph{local distance} $D_l$ computed from the matched time steps, which minimizes the distance of pairs with the same label and maximizes for pairs with different labels.


However, since the local distance only considers the difference between features at a \emph{single} matched time step, metric learning on the local distance does not capture \emph{temporal dependencies} appearing across consecutive time steps. Temporal dependency is known to greatly enhance the performance of time series classifications\,\cite{Hochreiter97}.
For example, when we try to distinguish between forwards and backwards from a set of motion images of a walking person, an image of a single time step gives very limited information on the direction, whereas a set of images of consecutive time steps provides clear clues. It is, therefore, more effective to distinguish samples with different labels by applying metric learning on the \emph{window distance} $D_w$ computed from consecutive time steps (e.g., $\{t_1, t_2, t_3\}$), as shown in Figure \ref{fig:motivation}(b). The existing algorithms miss out this opportunity as they focus on alignment at the expense of the temporal dependencies. This calls for a new method that can achieve both alignment and temporal dependencies. 

\begin{figure}[t!]
\includegraphics[width=7cm]{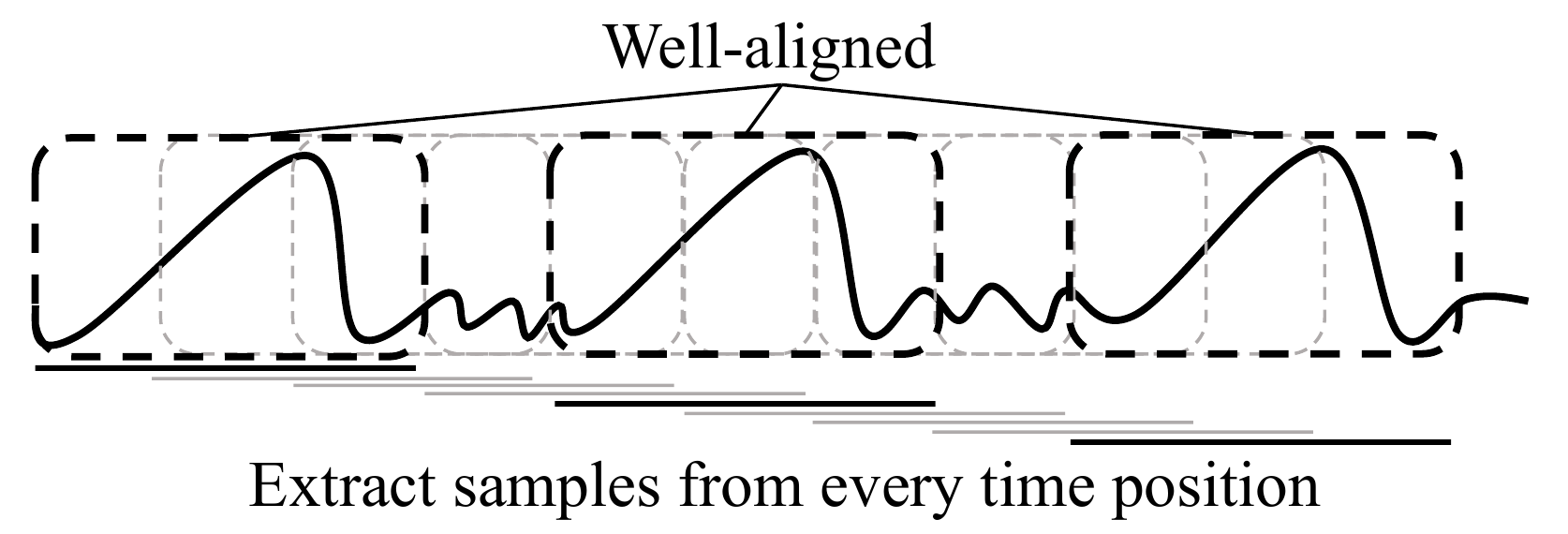}
\vspace*{-0.3cm}
\caption{Sliding window augmentation.}
\label{fig:intro_sliding_window_augmentation}
\vspace*{-0.7cm}
\end{figure}

In this paper, we propose a novel metric learning algorithm for kNN in streaming time series, called \emph{\textbf{\algname{}} (\underline{M}etric \underline{L}earning considering \underline{A}lignment and \underline{T}emporal dependencies)}. To account for both alignment and temporal dependency in learning a distance measure, \algname{} consists of the two main phases:
\begin{enumerate}[leftmargin=10pt, noitemsep, label={\arabic*.}]
\item {\bf Sliding Window Augmentation}: 
\algname{} augments a given time series using sliding window sampling\,\cite{Ye09} to achieve the alignment effect while preserving the temporal dependencies. As shown in Figure \ref{fig:intro_sliding_window_augmentation}, by extracting all possible samples as sets of consecutive time steps with a fixed length, the samples having similar patterns can be well-aligned even without explicit time step matching. At the same time, because consecutive time steps are well preserved in a sample, \algname{} can exploit the temporal dependencies between them.

\item {\bf Time-Invariant Metric Learning}: The temporal dependencies between features of augmented samples are \emph{time-invariant}, meaning that the dependency of two features in a fixed time difference is consistent regardless of their absolute temporal locations; i.e., the dependency of two features at time $t$ and $t+1$ is equal to that of the two features at time $t+1$ and $t+2$. Thus, \algname{} tries to fulfill this inherent characteristic in learning distance metric using the large margin nearest neighbor\,(LMNN)\,\cite{Lu15}, constraining a Mahalanobis matrix to be a \emph{Block Toeplitz}\,\cite{Gray06} structure.


\end{enumerate}
Extensive experiments on \emph{four} real-world streaming time series data sets indicate that \algname{} results in promising kNN performance.


\vspace*{-0.1cm}
\section{Related Work}
\label{sec:related_work}

We briefly review several existing studies to find better distance measures in time series data with respect to the following aspects: 
\emph{(i)} alignment to appropriately handle variation, and \emph{(ii)} metric learning to learn a better distance function.

\vspace*{-0.15cm}
\subsection{Alignment}
Dynamic time warping\,(DTW)\,\cite{Berndt94} is the most represensatice method to match the time steps of two samples in order to find the best warping path. Numerous variants of DTW have been studied, whose focus is on resolving the inefficiencies in time step matching\,\cite{Mueen16, Rakthanmanon12} and the invalidity of the triangle inequality in the learned distance\,\cite{Cuturi07, Hogeweg84}. 
However, DTW-based alignment methods fail to deal with the temporal dependencies because they explicitly match the time steps.

\vspace*{-0.15cm}
\subsection{Metric Learning}
Most time series metric learning algorithms in the literature apply metric learning to the local distance computed using DTW based alignment. LDML-DTW\,\cite{Mei15} and LMNN-DTW\,\cite{Shen17} first match time steps by multivariate dynamic time warping\,(MDTW)\,\cite{Berndt94}, and subsequently learn the Mahalanobis matrix of the local distance using LogDet divergence and Large Margin Nearest Neighbor\,(LMNN)\,\cite{Weinberger09}, respectively. DECADE\,\cite{Che17} devised a new alignment method to learn the valid distance measure while using deep networks to capture the complex dependencies in the local distance. There have been a few studies for learning the distance representation in the embedded space obtained from the last hidden layer of LSTM\,\cite{Mueller16}. However, to the best of our knowledge, no studies yet simultaneously consider alignment and temporal dependencies in time series metric learning.

\vspace*{-0.1cm}
\section{Preliminary}
\label{sec:preliminary}


\subsection{Problem Setting}

\label{sec:notation}
\begin{figure}[t!]
\includegraphics[width=7cm]{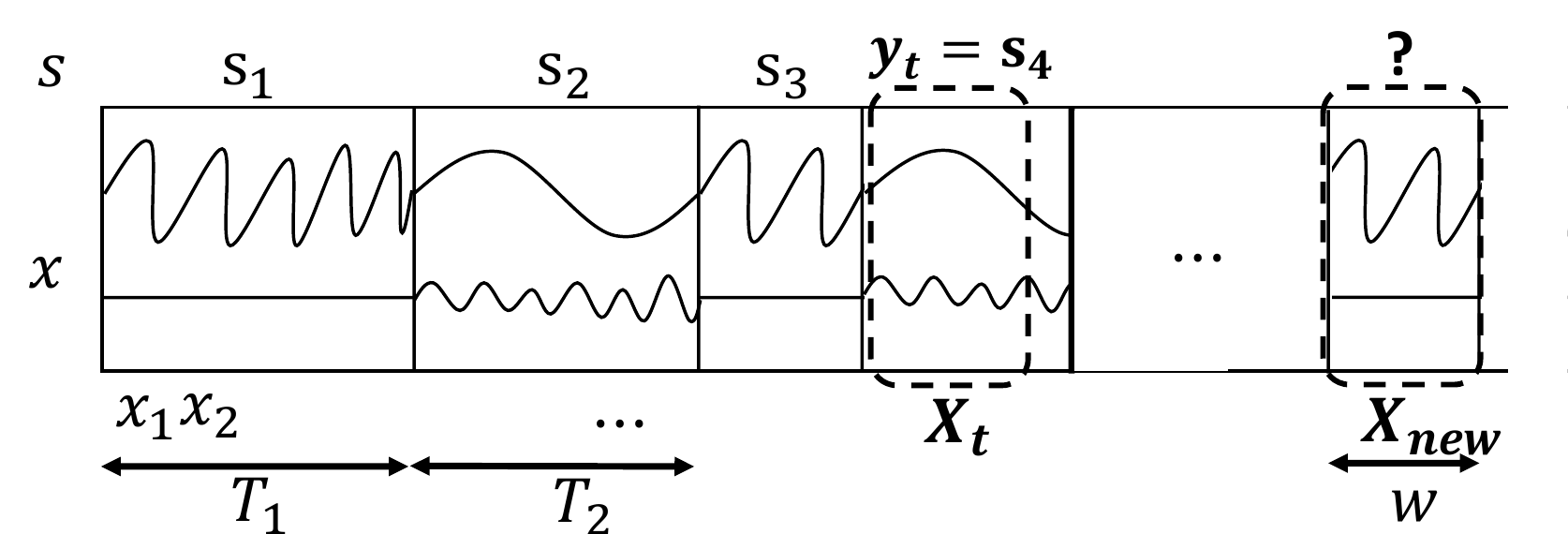}
\vspace*{-0.3cm}
\caption{Stream kNN settings.}
\label{fig:Stream_knn}
\vspace*{-0.5cm}
\end{figure}

We introduce the main concepts in the kNN classification problem on streaming time series in the following definitions, as also illustrated in Figure \ref{fig:Stream_knn}.

\begin{definition}
\label{def:multivariate_streaming_timeseries}
{({\sc Streaming time series})} A \emph{multivariate streaming time series} $x=[x_1,x_2,\cdots,x_{\mathbb{T}}]$ is a sequential observation of $x_i\in\mathbb{R}^d$ with $d$ features, where $\mathbb{T}$ is the length of $x$.
\end{definition}

\begin{definition}
\label{def:state}


{({\sc State})} A \emph{state} $s_j\in\{1,2,\cdots,S\}$ denotes the label of a sequence in a time series $x$ during the time period $T_{j}\,(\subset \mathbb{T})$, where $S$ is the number of states in $x$ and $j$ is the index of the state. The time series $x$ consists of various sizes of sequences with different states; for example, a time series $x$ collected from wearable devices for $15$ min consists of the two sequences: a "walk" state of the first $10$ min and a "run" state of the last $5$ min.

\end{definition}
\begin{definition}
\label{def:subsequence}
{({\sc Sample})} A \emph{sample} $X_t=[x_{t},\cdots,x_{t+w-1}]\in \mathbb{R}^{d\times w}$ is a subsequence of size $w$ extracted from a certain sequence in time period $T_{j}$ of a time series $x$. It consists of the consecutive time steps from time $t$ to time $t+w-1$. The state $s_j$ of the sequence becomes the label $y_t$ of $X_t$.
\end{definition}

\begin{definition}
\label{def:stream_knn}
{({\sc kNN classification problem})} Let $\mathbb{D}$ be a set of samples from a time series $x$. The kNN classification problem in streaming time series classifies the label of a sample $X_{new}$ by referring to that of the k nearest neighbors of $X_{new}$ in $\mathbb{D}$.
\end{definition}



\vspace*{-0.3cm}
\subsection{Large Margin Nearest Neighbor\,\cite{Weinberger09}}
\label{sec:lmnn}
Most metric learning methods aim at learning a Mahalanobis distance matrix $\mathbf{M}$\,\cite{Weinberger09, Davis07}, where the distance is defined as:


\vspace*{-0.2cm}
\begin{equation}
\small
D_{\mathbf{M}}(X_i,X_j)=(X_i-X_j)^T \mathbf{M} (X_i-X_j),
\label{eq:mahalanobis}
\end{equation}


where $X_i$ and $X_j\in \mathbb{R}^{dw}$ are two samples and $\mathbf{M} \in \mathbb{R}^{dw\times dw}$ is a positive semidefinite matrix.
Given a set of samples $\mathbb{D}$, for each sample $X_i \in \mathbb{D}$, LMNN first finds the $k$ nearest neighbors of $X_i$ based on the Euclidean distance and calls them \emph{target} neighbors. Subsequently, LMNN learns the optimal Mahalanobis distance matrix $\mathbf{M}$ which minimizes the following loss function:

\vspace*{-0.3cm}
\begin{equation}
\small
\begin{split}
\label{eq:cost_function}
\underset{\mathbf{M}\succeq 0}{\min}\ &\mathcal{E}^{lmnn}(\mathbf{M}) = (1-c)\sum_{ij} \eta_{ij} {D_\mathbf{M}}(X_i,X_j)\\
&+ c\sum_{ijl} \eta_{ij}(1-y_{il})\left[1+D_\mathbf{M}(X_i,X_j)+D_\mathbf{M}(X_i,X_l)	\right]_{+} ,
\end{split}
\end{equation}


where $c\in [0,1]$ controls the weights of the two penalizing terms, $\eta_{ij}\in \{0,1\}$ indicates whether $X_j$ is a target neighbor of $X_i$, $y_{il}\in\{0,1\}$ indicates whether $X_i$ and $X_l$ have the same label or not. Here, the first term penalizes the large distance from each sample to its $k$ target neighbors, and the second term penalizes the small distance from each sample to all other samples that do not share the same label.

\section{Methodology}
\label{sec:methodology}

\algname{} consists of two phases: (1) Sliding window augmentation, and (2) Time-invariant metric learning. Algorithm 1 describes the overall procedure of \algname{}.

\subsection{Phase I: Sliding Window Augmentation}
\label{sec:sliding_window_augmentation}

\begin{figure}[t!]
\centering
\begin{subfigure}[t!]{0.6\linewidth}
\includegraphics[width=\textwidth]{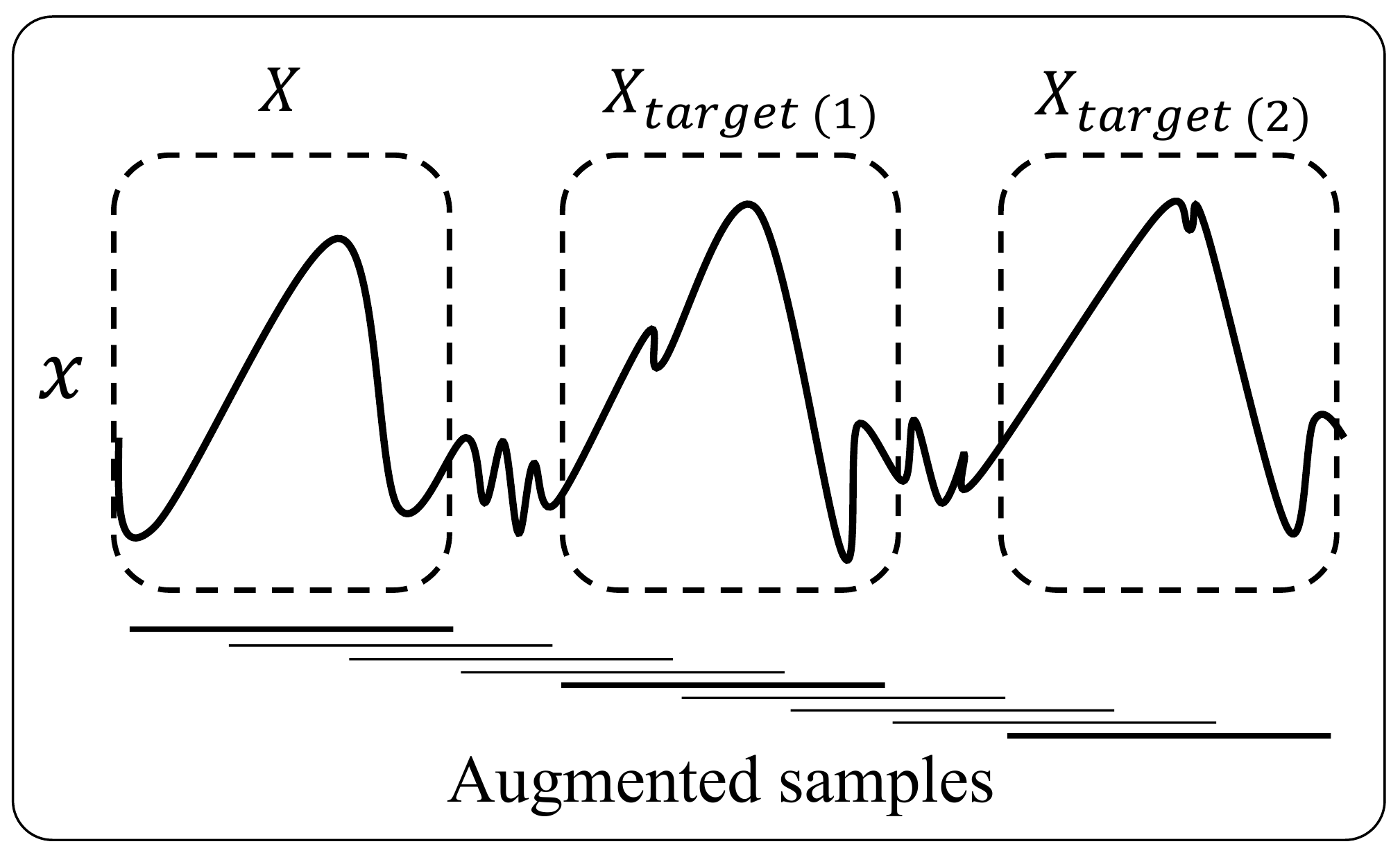}
\vspace*{-0.5cm}
\caption{Phase I.}
\label{fig:augmentation_effect}
 \end{subfigure} 
\begin{subfigure}[t!]{0.37\linewidth}
\includegraphics[width=\textwidth]{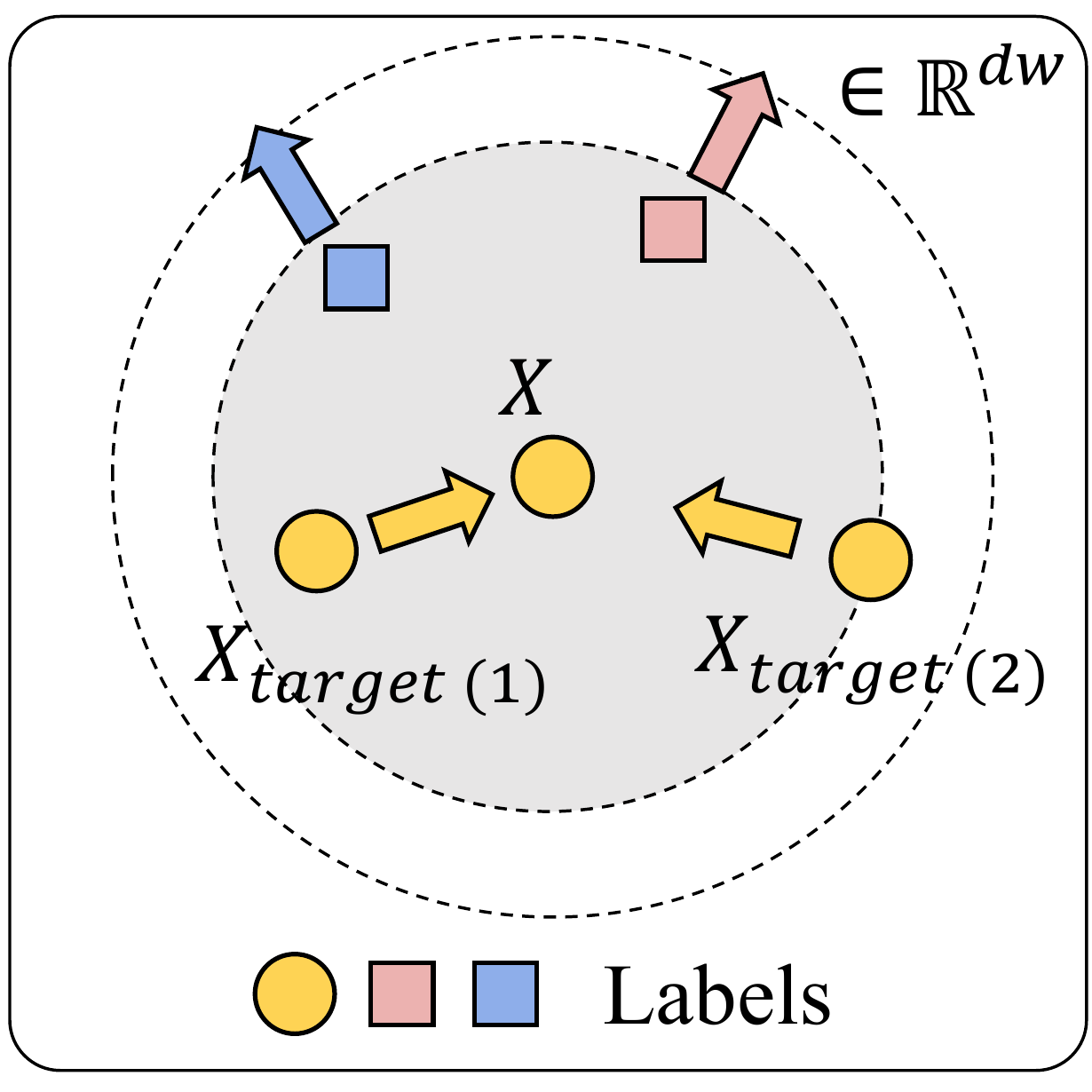}
\vspace*{-0.5cm}
\caption{Phase II.}
\label{fig:LMNN}
\end{subfigure}
\vspace*{-0.4cm}
\caption{Overall procedure of \algname{}.}
\label{fig:effect_of_augmentation_to_lmnn}
\vspace*{-0.3cm}
\end{figure}

Phase I enables metric learning to consider alignment and temporal dependencies concurrently. Specifically, all possible samples in $x$ can be extracted by sliding a fixed sized window from the beginning of $x$. Some parts of $x$ where the state switches are exempted from the augmentation.
These augmented samples become the training set for metric learning and help to achieve the alignment effect.
As in Figure \ref{fig:effect_of_augmentation_to_lmnn}(a), a sample $X$ has the most well-aligned training sample $X_{target(i)}$ as its neighbor with euclidean distance. Therefore, \algname{}, which is based on LMNN, exploits the well-aligned samples $X_{target(i)}$ as the target neighbor of $X$ if they share the same label and consequently learns the distance that makes them close as in Figure \ref{fig:effect_of_augmentation_to_lmnn}(b). Temporal dependencies can also be considered because metric learning is applied on window distance ($\mathbf{M} \in\mathbb{R}^{dw\times dw}$).

{
\newcommand{\INDSTATE}[1][1]{\STATE\hspace{#1\algorithmicindent}}
\setlength{\textfloatsep}{-0.7pt}
\begin{algorithm}[t!]
\caption{ \algname{}}
\label{alg:pseudocode}
\small
\begin{algorithmic}[1]
\REQUIRE A multivariate streaming time series $x$;
\ENSURE The optimal Mahalanobis matrix $M$;
\STATE \COMMENT{{\sc Phase I: sliding window augmentation}}
\STATE $X \leftarrow$ Extract augmented samples from $x$;
\STATE \COMMENT{{\sc Phase II: time-invariant metric learning}}
\STATE $\mathbf{L}\leftarrow I, \mathbf{M}^0\leftarrow I, \mathbf{Z}^0\leftarrow 0,\mathbf{U}^0\leftarrow 0$; \COMMENT{Initialize parameters.}
\INDSTATE[0] {\bf for} $k=1,2,\cdots,K$ (until convergence) \textbf{do} \COMMENT{Update parameters.}
\INDSTATE[1] {\bf repeat}
\INDSTATE[2] Calculate $\bigtriangledown \mathbf{M}^{k+1} \leftarrow {\partial\mathcal{E}^{lmnn}/ \partial\mathbf{M}^{k+1}}+\rho(\mathbf{M}^{k+1}-\mathbf{Z}^k+\mathbf{U}^k)$;
\INDSTATE[2] Calculate $\bigtriangledown \mathbf{L} \leftarrow 2\mathbf{L}\times{\bigtriangledown \mathbf{M}^{k+1}}$;
\INDSTATE[2] Update $\mathbf{L} \leftarrow (\mathbf{L}-\alpha\bigtriangledown\mathbf{L})$ and $\mathbf{M}^{k+1}\leftarrow \mathbf{L}^T\mathbf{L}$;
\INDSTATE[1] {\bf until} Converge;
\INDSTATE[1] $\mathbf{Z}^{k+1}_{B^{(m)}_{ij}} \leftarrow
 \sum_{l=1}^{R^{(m)}}{(\mathbf{M}^{k+1}+\mathbf{U}^{k})_{B^{(m)}_{ij,l}}/{R^{(m)}}}$;
\INDSTATE[1] $\mathbf{U}^{k+1} \leftarrow \mathbf{U}^{k}+(\mathbf{M}^{k+1}-\mathbf{Z}^{k+1})$;
\RETURN $\mathbf{M}^K$;
\end{algorithmic}
\end{algorithm}
}
\noindent

\subsection{Phase II: Time-Invariant Metric Learning}
\label{sec:time_invariant_metric_learning}

Phase II allows metric learning to capture the time-invariant property of the augmented samples, making it more effective for kNN.

\subsubsection{Block Toeplitz}
\label{sec:block_toeplitz}

As a result of Phase I, the consecutive samples are mostly overlapped. This leads to the covariance matrix of augmented samples in the form of \emph{Block Toeplitz}\,\cite{Gray06} as stated in Definition \ref{def:block_toeplitz_matrix}, which is proven by Theorem \ref{theorem:time_invariant_covariance}.




\renewcommand*{\arraystretch}{.4}


\begin{definition}
\label{def:block_toeplitz_matrix}
{({\sc Block toeplitz\,\cite{Gray06}})} A $dw\times dw$ \emph{block toeplitz} matrix $\mathbf{A}$ with $d\times d$ sub-blocks $\mathbf{A}^{(m)}$ has the following form:
\begin{equation}
\small
\begin{split}
\label{eq:block_toeplitz_matrix}
\mathbf{A} &=
\begin{bmatrix}
\mathbf{A}^{(0)} & (\mathbf{A}^{(1)})^T & (\mathbf{A}^{(2)})^T & \cdots & (\mathbf{A}^{(w-1)})^T \\
\mathbf{A}^{(1)} & \mathbf{A}^{(0)} & (\mathbf{A}^{(1)})^T & \ddots & \vdots \\
\mathbf{A}^{(2)} & \mathbf{A}^{(1)} & \ddots & (\mathbf{A}^{(1)})^T & (\mathbf{A}^{(2)})^T \\
\vdots & \ddots  & \mathbf{A}^{(1)} & \mathbf{A}^{(0)} & (\mathbf{A}^{(1)})^T \\
\mathbf{A}^{(w-1)} & \cdots & \mathbf{A}^{(2)} & \mathbf{A}^{(1)} & \mathbf{A}^{(0)} \\
\end{bmatrix}\ ,
\end{split}
\end{equation}
where the sub-block $\mathbf{A}^{(m)}$ appears $w-m$ times in $\mathbf{A}$ and has the same value at all occurrences.
\end{definition}


\begin{theorem}
\label{theorem:time_invariant_covariance}
Let $\mathbb{D}_{s}=[X_1, X_2, \cdots, X_L]\in \mathbb{R}^{dw\times L}$ be a set of samples obtained by sliding window augmentation, where $X_t \in \mathbb{R}^{dw}$ and $L\gg w$. 
The covariance matrix $\boldsymbol{\Sigma} \in \mathbb{R}^{dw\times dw}$ of $\mathbb{D}_{s}$ is in the form of Block Toeplitz.
\end{theorem}
\vspace*{-0.3cm}
\renewcommand*{\arraystretch}{.9}
\begin{proof}
$\mathbb{D}_{s}$ is represented with $w$ row partitions as follows:


\vspace*{-0.2cm}
\begin{equation}
\small
\begin{split}
\label{eq:sample_matrix}
\mathbb{D}_{s} &=
\begin{bmatrix}
x_1 & x_2 & x_3 & \cdots & x_{L-w+1} \\
x_2 & x_3 & x_4 & \cdots & x_{L-w+2} \\
\vdots & \vdots & \vdots &  & \vdots \\
x_w & x_{w+1} & x_{w+2} & \cdots & x_L \\
\end{bmatrix}\ ,
\end{split}
\end{equation}


where $x_i \in \mathbb{R}^{d}$. Since most portions in each row partition overlap, the mean $\mu_p\in \mathbb{R}^{d}$ of each row partition is approximately the same. Then, by the definition of the covariance matrix, $\boldsymbol{\Sigma} = 1/L \cdot (\mathbb{D}_{s}-\mu)(\mathbb{D}_{s}-\mu)^T$ approximately takes Block Toeplitz form, where $\mu=[\mu_p,\cdots,\mu_p]^T\cdot 1_L^T\in \mathbb{R}^{dw\times L}$.
\end{proof}
\renewcommand*{\arraystretch}{1.0}
The $(i, j)$-th element of the covariance matrix corresponds to the dependency between the $i$-th and $j$-th features, and if the matrix is Block Toeplitz, the dependency between the two features is time-invariant\,\cite{Hallac17}. Note that the time-invariant dependencies of the augmented samples should be preserved when learning distance.


\subsubsection{Time-Invariant Constraints}
\label{sec:time_invariant_constraints}
Here, we first analyze how the Mahalanobis matrix handles the dependency between two features. Eq. (\ref{eq:mahalanobis}) can be represented as follows:


\vspace*{-0.2cm}
\begin{equation}
\small
\begin{split}
\label{eq:mahalanobis_distance}
D_{\mathbf{M}}(X_i,X_j) &= tr(\mathbf{M}(X_1-X_2)^T(X_1-X_2))\\
&= \sum_{ij} \mathbf{M}_{ij}(X_1^{(i)}-X_2^{(i)})(X_1^{(j)}-X_2^{(j)}),
\end{split}
\end{equation}
where $X_1^{(i)}, i\in \{1,\cdots,dw\}$, is the $i$-th feature scala value of a sample $X_1$, and $\mathbf{M}_{ij}$ is the $(i,j)$-th element of $\mathbf{M}$. $\mathbf{M}_{ij}$ decides how much the product of the distance between $i$-th features and that of $j$-th features affects the total distance, meaning that $\mathbf{M}_{ij}$ handles the dependency between the $i$-th and $j$-th features.

To preserve the time-invariant dependencies, we propose the following time-invariant metric learning framework:


\vspace*{-0.2cm}
\begin{equation}
\small
\begin{split}
\label{eq:time_invariant_metric_learning}
\underset{\mathbf{M}}{\min} \quad \mathcal{E}^{lmnn}(\mathbf{M}) \quad \qquad\\
\rm{\text{subject to}}:\ \mathbf{M} \succeq 0,\ \mathbf{M} \in \mathcal{T},
\end{split}
\end{equation}
where $\mathcal{T}$ is a set of $dw\times dw$ symmetric Block Toeplitz matrices. By constraining $\mathbf{M}$ to be Block Toeplitz, the distances between two features in the same time difference should be regarded equally, with the same values as in $\mathbf{M}$. Thus, the optimal matrix can effectively capture the time-invariant dependencies structure.



\subsubsection{Optimization}
\label{sec:optimization}
We solve the proposed optimization problem using the alternating direction method of multipliers (ADMM)\,\cite{Boyd11}. We transform Eq.\,(\ref{eq:time_invariant_metric_learning}) into an ADMM-friendly form as follows:


\vspace*{-0.2cm}
\begin{equation}
\small
\begin{split}
\label{eq:admm_form}
\mathcal{L}_{\rho}(\mathbf{M},&\mathbf{Z},\mathbf{U}) = \mathcal{E}^{lmnn}(\mathbf{M})+{\rho \over 2}\parallel \mathbf{M}-\mathbf{Z}+\mathbf{U} \parallel_F^2 \\
&\rm{\text{subject to}}: \mathbf{Z}\in \mathcal{T},\ \mathbf{Z}=\mathbf{M},\ \mathbf{M} \succeq 0.
\end{split}
\end{equation}


Then, the following three steps are repeated until convergence.


\vspace*{-0.2cm}
\renewcommand*{\arraystretch}{0}
\begin{equation}
\small
\begin{split}
\label{eq:admm_iteration}
(a)\ \mathbf{M}^{k+1} &= \underset{\mathbf{M}}{argmin}\ \mathcal{L}_{\rho}(\mathbf{M},\mathbf{Z}^k,\mathbf{U}^k)\\
(b)\ \mathbf{Z}^{k+1} &= \underset{\mathbf{Z}}{argmin}\ \mathcal{L}_{\rho}(\mathbf{M}^{k+1},\mathbf{Z},\mathbf{U}^k)\\
(c)\ \mathbf{U}^{k+1} &= \mathbf{U}^{k}+(\mathbf{M}^{k+1}-\mathbf{Z}^{k+1}).
\end{split}
\end{equation}


\renewcommand*{\arraystretch}{1.0}

\noindent\textbf{\uline{$\mathbf{M}$-update:}} The $\mathbf{M}$-update can be written as:


\begin{equation}
\small
\begin{split}
\label{eq:m_update}
\mathbf{M}^{k+1} &= \underset{\mathbf{M}\succeq 0}{argmin}\ \mathcal{E}^{lmnn}(\mathbf{M})+{\rho \over 2}\parallel \mathbf{M}-\mathbf{Z^k}+\mathbf{U^k} \parallel_F^2.
\end{split}
\end{equation}


This problem can be solved by the gradient descent method. To ensure a positive semi-definite $\mathbf{M}$, \algname{} factorizes $\mathbf{M}$ as $\mathbf{M}=\mathbf{L}^T\mathbf{L}$ and updates $\mathbf{L}$ through sub-gradient descent. Here, the sub-gradient of Eq. (\ref{eq:m_update}) with respect to $\mathbf{L}$ is ${\partial\mathcal{E}^{lmnn}\over\partial\mathbf{L}}=2\mathbf{L}({\partial\mathcal{E}^{lmnn}\over\partial\mathbf{M}}+\rho(\mathbf{M}-\mathbf{Z}^k+\mathbf{U}^k))$ by the chain rule. 

\vspace*{0.1cm}
\noindent\textbf{\uline{$\mathbf{Z}$-update:}}
The closed form solution of the $\mathbf{Z}$-update is as follows:

\vspace*{-0.4cm}
\begin{equation}
\small
\begin{split}
\label{eq:closed_form}
\mathbf{Z}^{k+1}_{B^{(m)}_{ij}} &= {\sum_{l=1}^{R^{(m)}}{(\mathbf{M}^{k+1}+\mathbf{U}^{k})_{B^{(m)}_{ij,l}} / {R^{(m)}}}},
\end{split}
\end{equation}
where $B^{(m)}_{ij,l}$ is the index $(x, y)$ at $\mathbf{Z}$ of the $(i, j)$-th element of the $l$-th occurrence sub-matrix $\mathbf{Z}^{(m)}$, and $R^{(m)}$ is the number of occurrences of $\mathbf{Z}^{(m)}$ in $\mathbf{Z}$. Refer to\,\cite{Hallac17} for details.

\section{Experiments}
\label{sec:experiments}
To validate the superiority of \algname{}, we performed the kNN classification task on \emph{four} real-world streaming time series data sets. The experimental results confirmed that \algname{} maintains its dominance over other algorithms.

\subsection{Experiment Setup}
\label{sec:experiment_setup}
\subsubsection{Data Sets}
\label{sec:datasets}


\begin{table}[t!]
\small
\caption{Summary statistics of the UCI dataset.}
\vspace*{-0.4cm}
\label{table:dataset}
\begin{tabular}{|c|c|c|c|}
\hline
Data Set & Description                                                   & \ Dims & \ $\#$ Lables \\ \hline
SCMA\,\cite{SCMA}     & 3D wearable accelerometer sensors       & 3                & 7            \\ \hline
AReM\,\cite{AReM}     & Wireless Sensor Network    & 6                & 6            \\ \hline
EMG\,\cite{EMG}      & Eight Electromyogram(EMG) sensors & 8                & 5        \\ \hline
Vicon\,\cite{Vicon}    & Nine Vicon 3D trackers    & 27               & 5        \\ \hline
\end{tabular}
\vspace*{-0.3cm}
\end{table}


The statistics of the four benchmark data sets are summarized in Table \ref{table:dataset}. Note that we only used $5$ ambiguous labels defined as normal activities in EMG and Vicon data sets for a more difficult kNN task.


\subsubsection{Algorithms}
\label{sec:compared_methods}
We compared \algname{} with \emph{four} existing algorithms for measuring distance in time series:
\begin{itemize}[leftmargin=12pt, noitemsep]
\item ED : Euclidean distance (baseline).
\item MDTW\,\cite{Berndt94} : Multivariate dynamic time warping.
\item LDMLT\,\cite{Mei15} : Learning local distance by DTW alignment.
\item LMNN\,\cite{Lu15} : Large Margin Nearest Neighbor.
\item {\bf \algname{}} : Our proposed method.
\end{itemize}

\subsubsection{Settings}
\label{sec:settings}

We assumed that $510$ consecutive observations ($|T_{j}|$ = $510$) from each label are given for each data set, except AReM where $|T_{j}|$ is $480$. 
The training samples, where the size $w$ is set to 10, were extracted from the given observations.
For existing algorithms, we randomly extracted 100 samples from each label.
We set $k$ to 1 for kNN because 1-NN is widely accepted as the most accurate for many tasks\,\cite{Ding08}.
We randomly selected the starting time of the consecutive observations and reported the average results of the five experiments for each dataset.

\subsection{Experimental Results}
\label{sec:experiment_results}


\begin{table}[t!]
\caption{Accuracy\,(\%) of kNN with standard deviation ($k=1$).}
\vspace*{-0.5cm}
\begin{center}
{\small
\begin{tabular}{|c|c|c|c|c|c|}\hline
{Data Set}&ED&DTW&LDMLT&LMNN&{\bf \algname{} }\\\hline\hline
{SCMA}& 97.5$\pm$0.6 & 98.5$\pm$0.5 & 97.4$\pm$0.5 & 99.1$\pm$0.4 &{\bf 99.5$\pm$0.1}\\\hline
{AReM}& 75.3$\pm$1.2 & 76.4$\pm$1.9 & {\bf 84.4$\pm$2.1} & 73.3$\pm$1.5 & 80.8$\pm$0.8\\\hline
{EMG}& 61.8$\pm$2.3 & 65.4$\pm$1.5 & 65.2$\pm$2.8 & 65.9$\pm$2.1 & {\bf 69.3$\pm$1.7}\\\hline
{Vicon}&73.3$\pm$1.9 & 75.8$\pm$2.1 & 72.2$\pm$3.3 & 81.5$\pm$2.6 & {\bf 84.9$\pm$1.8}\\\hline
\end{tabular}
}
\end{center}
\label{table:accuracy}
\vspace*{-0.4cm}
\end{table}


\subsubsection{Overall Accuracy}
\label{sec:overall_accuracy}
Table \ref{table:accuracy} shows the kNN accuracy of the five algorithms. In the SCMA, EMG, and Vicon data sets,  \algname{} achieved the highest accuracy compared with other algorithms. LDMLT produced the highest accuracy for the AReM data set, but \algname{} also achieved a significant improvement compared with the remaining algorithms. This emphasizes the need to consider both alignment and temporal dependencies.


\begin{table}[t!]
\caption{Accuracy improvement\,(\%) on augmented samples.}
\vspace*{-0.5cm}
\begin{center}
{\small
\begin{tabular}{|c|c|c|c|c|}\hline
{Data Set}&ED&DTW&LDMLT&LMNN\\\hline\hline
{SCMA}& 1.02 & 0.40 & 1.33 & 0.20 \\\hline
{AReM}& -0.21 & 0.52 & 1.30 & 0.41\\\hline
{EMG}& 5.34 & 3.52 & -1.4 & 2.43\\\hline
{Vicon}& 3.27 & -0.13 & -2.2 & 2.33\\\hline
\end{tabular}
}
\end{center}
\label{table:phase1_effect}
\vspace*{-0.4cm}
\end{table}


\subsubsection{Effect of Phases I and II}
\label{sec:effect_of_phases}
All the existing algorithms were also evaluated on augmented samples obtained in Phase I. As shown in Table \ref{table:phase1_effect}, the accuracy of the algorithms were significantly improved by up to $5.3\%$ compared with the accuracy from randomly extracted samples. This shows that augmenting well-aligned samples is evidently beneficial for the time series classification.

In addition, \algname{} still outperformed LMNN by up to $9.8\%$, though LMNN performed on augmented samples. The performance of metric learning for time series can thus be further boosted by preserving the time-invariant characteristic.


\section{Conclusion}
\label{sec:conclusion}
In this paper, we proposed \algname{}, a novel metric learning algorithm that considers the two main characteristics in time series, i.e., variation and temporal dependencies, by using sliding window augmentation and time-invariant metric learning, respectively. Using four real-world data sets, we showed that \algname{} outperforms the existing algorithms in most cases. For future work, we will tackle more challenging settings where more complex variations and temporal dependencies exist.


\begin{acks}
This work was partly supported by the National Research Foundation of Korea\,(NRF) grant funded by the Korea government\,(Ministry of Science and ICT) (No.\ 2017R1E1A1A01075927) and the MOLIT (The Ministry of Land, Infrastructure and Transport), Korea, under the national spatial information research program supervised by the KAIA\,(Korea Agency for Infrastructure Technology Advancement) (19NSIP-B081011-06).
\end{acks}
\medskip

\balance

\bibliographystyle{ACM-Reference-Format}
\bibliography{7-reference} 

\end{document}